\documentclass[11pt,letterpaper]{article}
\usepackage{ijcnlp2017}
\usepackage{times}
\usepackage{latexsym}
\usepackage{multirow}
\usepackage{graphicx}

\ijcnlpfinalcopy

\def\ijcnlppaperid{***}

\title{OhioState at IJCNLP-2017 Task 4: Exploring Neural Architectures for Multilingual Customer Feedback Analysis}

\author{Dushyanta Dhyani \\
 {The Ohio State University OH, USA}\\
  {\tt dhyani.2@osu.edu}}

\date{}

\begin{document}

\maketitle

\begin{abstract}
This paper describes our systems for IJCNLP 2017 Shared Task on Customer Feedback Analysis. We experimented with simple neural architectures that gave competitive performance on certain tasks. This includes shallow CNN and Bi-Directional LSTM architectures with Facebook's Fasttext as a baseline model. Our best performing model was in the Top 5  systems using the Exact-Accuracy and Micro-Average-F1 metrics for the Spanish (85.28\% for both) and French (70\% and 73.17\% respectively) task, and outperformed all the other models on comment (87.28\%) and meaningless (51.85\%) tags using Micro Average F1 by Tags metric  for the French task. 
 
\end{abstract}

\section{Introduction}

Customer Feedback Analysis (CFA) aims to analyze the feedback given by customers to various products/organizations. A primary component of CFA is to identify what the feedback is discussing so that further processing can be carried out appropriately. This requirement serves as a motivation for this shared task, which aims to classify user feedback in multiple languages into pre-defined categories and automate the process using machine learning methods for document classification. 

\section{Related Work}

Document Classification is a well-studied problem in the NLP community with applications like sentiment analysis \cite{chen2016neural}, language identification \cite{lui2012langid}, email/document routing and even adverse drug reaction classification \cite{huynh2016adverse}. However, the problem and various proposed solutions are highly domain and use-case specific. State of the art sentiment analysis models/architectures that perform well for news articles fail to perform well for Twitter sentiment analysis. Moreover, the Twitter sentiment analysis models have to be re-designed for tasks like target dependent sentiment analysis \cite{vo2015target}. This shows that the type of models used for a particular domain depends a lot on the data and the granularity of the categories. Recent efforts \cite{kim2014convolutional,zhang2015character,conneau2016very,yang2016hierarchical,joulin2016bag} show the applicability of a single (generally neural) model over a variety of datasets, showing their capability to model text classification tasks in a domain and language agnostic way.

\section{Task Description}

The goal of the shared task is: Given customer feedback in four languages (English, French, Spanish and Japanese) the participants should design systems that can classify customer feedback into pre-defined categories (comment, request, bug, complaint, meaningless and undetermined). Evaluation is done on per-language basis using a variety of metrics. 

\subsection{Dataset}

The contest organizers provided customer feedback data in four languages. The size of train/dev/test samples for each of the sub task is shown in Table ~\ref{table:datasizestats}. About 5\% of the samples across the data splits for English, French and Japanese task have multiple labels, while each sample in the Spanish task has only one label. For the data samples with a single label, the distribution was highly biased towards certain classes, with the comment and complaint categories covering 80\%-95\% across all data splits for each sub-task. The contest organizers also provided a relatively larger corpus of unlabeled data. While this could be used in different ways like learning domain-specific word embeddings, we exclude it in our experiments.

\begin{table}[!ht]
\centering
\begin{tabular}{|c| c |c|c|} 
 \hline
 \textbf{Language} & \textbf{Train} & \textbf{Dev} & \textbf{Test} \\ [0.5ex] 
 \hline
 \textbf{English(en)} & 3066 & 501 & 501 \\
\textbf{ Spanish(es)} & 1632 & 302 & 300 \\
\textbf{ French(fr)} & 1951 & 401 & 401 \\
\textbf{Japanese(jp)} &  1527 & 251 & 301 \\
 \hline
Total &  8176 & 1455 & 1503 \\ [1ex] 
 \hline
\end{tabular}
\caption{Number of Training Samples for each sub-task}
\label{table:datasizestats}
\end{table}

\subsection{Evaluation}

The contest organizers use 3 metrics to evaluate the submitted systems
\begin{itemize}
\item \textbf{Exact Accuracy}: All tags should be predicted correctly.
\item \textbf{Micro-Average F1}: As discussed in \cite{Manning:2008:IIR:1394399}, micro-average F1 gathers document level decisions across classes, thus giving more weight to large classes, which is the case across all the sub-tasks
\item \textbf{Micro-Average by Tags}: Label specific micro-average F1.
\end{itemize}

\begin{table*}[t]
\centering
\begin{tabular}{|c|c|c|c|c|c|} 
 \hline
 \multirow{2}{*}{} & \multicolumn{5}{|c|}{ \textbf{Sub-Tasks}} \\\cline{2-6}
 {} & \multicolumn{2}{|c|}{ \textbf{EN}} &  \textbf{ES} & \multicolumn{2}{|c|}{ \textbf{FR}} \\
  \hline
  \textbf{Models} & \textbf{EA} & \textbf{MAF} & \textbf{Both EA \& MAF} & \textbf{EA} & \textbf{MAF}\\
    \hline
 \textbf{OhioState-FastText} 	& 		\textbf{63.4} 		& 	\textbf{65.36}	 &  	82.94 	&  68  	&   70.49\\
\textbf{ OhioState-CNN} & 		54.20 		& 	56.13	 &  	81.27  &  65  	&   67.8\\
\textbf{ OhioState-biLSTM1}			& 61.2  		& 	63.79	 &  	82.61  &  \textbf{70 } 	&   \textbf{73.17}\\
\textbf{OhioState-biLSTM2} 		& 61.6   		& 	63.98	 &  	\textbf{85.28} &  68.5  	&   71.71\\
\textbf{OhioState-biLSTM3} 		& 62.8  		& 	64.97	 &  	79.93 &  65  	&   67.56\\
 \hline
\end{tabular}
\caption{Performance of Various Models for Exact Accuracy (EA) and Micro-Average F1 (MAF) score}
\label{table:modelperformace}
\end{table*}

\begin{table*}[]
\centering
\normalsize
\begin{tabular}{|c|c|c|c|c|c|}
\hline
\multicolumn{2}{|c|}{\textbf{ES}} & \multicolumn{4}{|c|}{\textbf{FR}} \\ 
\hline
 \multicolumn{2}{|c|}{\textbf{Both EA \& MAF}} & \multicolumn{2}{|c|}{\textbf{EA}} &
 \multicolumn{2}{|c|}{\textbf{MAF}} \\ 
 \hline
Plank-multilingual	&		88.63		&	Plank-monolingual	& 73.75 		&	Plank-monolingual			&	76.59	\\
Plank-monolingual	&		88.29	&		IITP-CNN-entrans		&	71.75		&		IITP-CNN-entrans	&		74.63	\\
IIIT-H-biLSTM&			86.29	&	Plank-multilingual	&	71.50		&		Plank-multilingual	&		74.39	\\
IITP-RNN	&		85.62	&		\textbf{OhioState-biLSTM1}	&	70.00		&	\textbf{OhioState-biLSTM1}		&		73.17	\\
\textbf{OhioState-biLSTM2}	&		85.28	&		IIIT-H-SVM	&	69.75 		&			ADAPT-Run1		&		72.68	\\
 \hline
\end{tabular}
\caption{Top-5 Performing systems for the Spanish and French Sub-Tasks}
\label{table:topsystems}
\end{table*}

\begin{table*}[]
\centering
\begin{tabular}{
|c|c|c|c|c|c|
}
\hline
\textbf{Task} & \textbf{Comments} & \textbf{Complaint} & \textbf{Meaningless} & \textbf{Bug} & \textbf{Request} \\
\hline
 \textbf{EN} & 	BiLSTM2 (77.8)	&	 BiLSTM1 (63.4)	&	 fastText (48.3)	& 	fastText (16.7)	&	 fastText (53.9)		\\
 \textbf{FR} & 	\textbf{BiLSTM2 (87.3)}	& 	BiLSTM1 (57.4)	& 	 \textbf{BiLSTM1 (51.9)}	& fastText (20)& fastText (15.4)				\\
 \textbf{ES} & 	BiLSTM2 (92.6)	& 	BiLSTM2 (68.9) 	& 0					& 		0	& fastText (31.6)				\\
 \hline
\end{tabular}
\caption{Our best performing models (F1) for each label of the English, French and Spanish sub-task (Scores in bold perform best amongst all submitted systems) } 
\label{table:bestsystemlabel}
\end{table*}


\section{Proposed Approach}
Motivated by the success of a variety of architectures for document classification task, we use multiple methods for the given challenge. We used a recently released open source tool called Fasttext as our baseline. In addition to that, we used a commonly used CNN architecture and multiple LSTM based architectures. In this section we discuss various components of our systems.
\subsection{Pre-processing}
We used minimal text pre-processing by using in-built tokenizer's from TensorFlow \cite{tensorflow2015-whitepaper} and Keras \cite{chollet2015keras} across all our architectures. In addition to that, we applied some elementary text cleaning to the English data only, given our lack of understanding of other languages. 

\subsection{Models}
\subsubsection{fastText (OhioState-FastText)}
Given its ease of use, we used the fastText \cite{joulin2016bag}  tool\footnote{We used the Python wrapper from pypi}  as our baseline model. At its core, fastText is a linear model with a few neat tricks to make the training fast and efficient. It takes individual word representations and averages them to get the representation of the given text. This representation is then passed through a softmax to get class distribution. Training is performed using Stochastic Gradient Descent to minimize the negative log-likelihood over all the classes.  We used most of the default parameters as in the original tool. We, however, found that the model performs best on the dev set when the embedding dimension is set to 200 and the model is trained for 100 epochs. Since the size of training data and number of training labels were small, we used the softmax loss function (and not the hierarchical softmax and negative sampling methods) as training time was not a constraint.

\subsection{Convolution Neural Networks (OhioState-CNN)}
We also performed some basic experiments with CNN's given their applicability to text classification \cite{kim2014convolutional,zhang2015character,conneau2016very,kalchbrenner2014convolutional} problems. We used a simplified version of the architecture from \cite{kim2014convolutional} as discussed here\footnote{http://www.wildml.com/2015/12/implementing-a-cnn-for-text-classification-in-tensorflow/}. We set the word embedding size to 100 and trained the architecture for 10 epochs (after which it starts overfitting) . We used 128 filters of filter width 3,4 and 5 and added a dropout layer with retention probability of 0.5. We trained the model using Adam \cite{kingma2014adam} and the sigmoid cross entropy loss.

\subsection{Bi-Directional LSTM}
LSTM's have been shown to be extremely effective for learning representations for text, not only for sequence to sequence labeling tasks, but for general classification tasks \cite{yang2016hierarchical} as well as language modeling \cite{li2015hierarchical}. We use Keras' ability to plug and play layers to experiment with a couple of architectures.
\begin{itemize}
\item \textbf{OhioState-biLSTM1} : A single layer Bi-directional LSTM with an embedding layer for the vocabulary and a dense layer with a sigmoid activation for the class labels. We also added a Dropout layer (with retention probability of 0.3) after the LSTM layer.
\item \textbf{OhioState-biLSTM2} : We added a 1D Convolutional (with ReLU activation) and Max-Pooling layer after the word embedding layer which has shown to better represent n-gram like characteristics in text.
\item \textbf{OhioState-biLSTM3} : We also added a  Batch Normalization layer after the Convolution layer in the above architecture(though it decreased the performance)
\end{itemize}

Note that we did not make use of any pre-trained embeddings.
We used the same training parameters for the 3 Bi-LSTM variants discussed above: Word embedding dimension, LSTM unit size and Batch Size were set to 64. We used the Adam \cite{kingma2014adam} optimizer with binary cross entropy loss.

A few points worth mentioning: While the CNN and Bi-LSTM architectures were trained in a multi-label setting, at prediction time, we only predict the label with the maximum score. Also, Japanese text in the corpus either has no or a single space and thus tokenization is not effective. So even though we achieve some (unconvincing) results for the Japanese task, we do not consider them as relevant to this sub-task which requires more sub-word level treatment.

\section{Results}

We report the performance on 3 sub-tasks (leaving out Japanese for reasons previously discussed) for our models  in Table ~\ref{table:modelperformace} and comparison with systems designed by other teams in Table ~\ref{table:topsystems} using the exact accuracy and micro-average F1 metric.

While there is a considerable difference between our best performing system and the top systems for the English sub-task, we obtain competitive performance for the Spanish and French sub-tasks. Moreover, our LSTM based models outperform other systems for \textbf{comment and meaningless} category when evaluated using Micro Average by Tags metric for the French sub-task with an F-1 accuracy of 87.28\% and 51.85\% respectively. However, as shown in Table ~\ref{table:bestsystemlabel},  our neural models failed to generalize to the infrequent labels as compared to a shallow model like fastText  which is an expected behavior.

\section{Conclusion}
We propose some simple but effective neural architectures for customer feedback analysis. We show the effectiveness of LSTM based models for Text Classification in French and Spanish sub-tasks without any prior information like heavy pre-trained embeddings, thus making it easy to perform fast and effective hyper-parameter tuning and architecture exploration.

\bibliography{ijcnlp2017}

\begin{thebibliography}{16}
\expandafter\ifx\csname natexlab\endcsname\relax\def\natexlab#1{#1}\fi

\bibitem[{Abadi et~al.(2015)Abadi, Agarwal, Barham, Brevdo, Chen, Citro,
  Corrado, Davis, Dean, Devin, Ghemawat, Goodfellow, Harp, Irving, Isard, Jia,
  Jozefowicz, Kaiser, Kudlur, Levenberg, Man\'{e}, Monga, Moore, Murray, Olah,
  Schuster, Shlens, Steiner, Sutskever, Talwar, Tucker, Vanhoucke, Vasudevan,
  Vi\'{e}gas, Vinyals, Warden, Wattenberg, Wicke, Yu, and
  Zheng}]{tensorflow2015-whitepaper}
Mart\'{\i}n Abadi, Ashish Agarwal, Paul Barham, Eugene Brevdo, Zhifeng Chen,
  Craig Citro, Greg~S. Corrado, Andy Davis, Jeffrey Dean, Matthieu Devin,
  Sanjay Ghemawat, Ian Goodfellow, Andrew Harp, Geoffrey Irving, Michael Isard,
  Yangqing Jia, Rafal Jozefowicz, Lukasz Kaiser, Manjunath Kudlur, Josh
  Levenberg, Dan Man\'{e}, Rajat Monga, Sherry Moore, Derek Murray, Chris Olah,
  Mike Schuster, Jonathon Shlens, Benoit Steiner, Ilya Sutskever, Kunal Talwar,
  Paul Tucker, Vincent Vanhoucke, Vijay Vasudevan, Fernanda Vi\'{e}gas, Oriol
  Vinyals, Pete Warden, Martin Wattenberg, Martin Wicke, Yuan Yu, and Xiaoqiang
  Zheng. 2015.
\newblock \href {https://www.tensorflow.org/} {{TensorFlow}: Large-scale
  machine learning on heterogeneous systems}.
\newblock Software available from tensorflow.org.

\bibitem[{Chen et~al.(2016)Chen, Sun, Tu, Lin, and Liu}]{chen2016neural}
Huimin Chen, Maosong Sun, Cunchao Tu, Yankai Lin, and Zhiyuan Liu. 2016.
\newblock Neural sentiment classification with user and product attention.
\newblock In \emph{EMNLP}, pages 1650--1659.

\bibitem[{Chollet et~al.(2015)}]{chollet2015keras}
Fran\c{c}ois Chollet et~al. 2015.
\newblock Keras.
\newblock \url{https://github.com/fchollet/keras}.

\bibitem[{Conneau et~al.(2016)Conneau, Schwenk, Barrault, and
  Lecun}]{conneau2016very}
Alexis Conneau, Holger Schwenk, Lo{\"\i}c Barrault, and Yann Lecun. 2016.
\newblock Very deep convolutional networks for natural language processing.
\newblock \emph{arXiv preprint arXiv:1606.01781}.

\bibitem[{Huynh et~al.(2016)Huynh, He, Willis, and
  R{\"u}ger}]{huynh2016adverse}
Trung Huynh, Yulan He, Alistair Willis, and Stefan R{\"u}ger. 2016.
\newblock Adverse drug reaction classification with deep neural networks.
\newblock In \emph{Proceedings of COLING 2016, the 26th International
  Conference on Computational Linguistics: Technical Papers}, pages 877--887.

\bibitem[{Joulin et~al.(2016)Joulin, Grave, Bojanowski, and
  Mikolov}]{joulin2016bag}
Armand Joulin, Edouard Grave, Piotr Bojanowski, and Tomas Mikolov. 2016.
\newblock Bag of tricks for efficient text classification.
\newblock \emph{arXiv preprint arXiv:1607.01759}.

\bibitem[{Kalchbrenner et~al.(2014)Kalchbrenner, Grefenstette, and
  Blunsom}]{kalchbrenner2014convolutional}
Nal Kalchbrenner, Edward Grefenstette, and Phil Blunsom. 2014.
\newblock A convolutional neural network for modelling sentences.
\newblock \emph{arXiv preprint arXiv:1404.2188}.

\bibitem[{Kim(2014)}]{kim2014convolutional}
Yoon Kim. 2014.
\newblock Convolutional neural networks for sentence classification.
\newblock \emph{arXiv preprint arXiv:1408.5882}.

\bibitem[{Kingma and Ba(2014)}]{kingma2014adam}
Diederik Kingma and Jimmy Ba. 2014.
\newblock Adam: A method for stochastic optimization.
\newblock \emph{arXiv preprint arXiv:1412.6980}.

\bibitem[{Li et~al.(2015)Li, Luong, and Jurafsky}]{li2015hierarchical}
Jiwei Li, Minh-Thang Luong, and Dan Jurafsky. 2015.
\newblock A hierarchical neural autoencoder for paragraphs and documents.
\newblock \emph{arXiv preprint arXiv:1506.01057}.

\bibitem[{Lui and Baldwin(2012)}]{lui2012langid}
Marco Lui and Timothy Baldwin. 2012.
\newblock langid. py: An off-the-shelf language identification tool.
\newblock In \emph{Proceedings of the ACL 2012 system demonstrations}, pages
  25--30. Association for Computational Linguistics.

\bibitem[{Manning et~al.(2008)Manning, Raghavan, and
  Sch\"{u}tze}]{Manning:2008:IIR:1394399}
Christopher~D. Manning, Prabhakar Raghavan, and Hinrich Sch\"{u}tze. 2008.
\newblock \emph{Introduction to Information Retrieval}.
\newblock Cambridge University Press, New York, NY, USA.

\bibitem[{Srivastava et~al.(2014)Srivastava, Hinton, Krizhevsky, Sutskever, and
  Salakhutdinov}]{srivastava2014dropout}
Nitish Srivastava, Geoffrey~E Hinton, Alex Krizhevsky, Ilya Sutskever, and
  Ruslan Salakhutdinov. 2014.
\newblock Dropout: a simple way to prevent neural networks from overfitting.
\newblock \emph{Journal of machine learning research}, 15(1):1929--1958.

\bibitem[{Vo and Zhang(2015)}]{vo2015target}
Duy-Tin Vo and Yue Zhang. 2015.
\newblock Target-dependent twitter sentiment classification with rich automatic
  features.
\newblock In \emph{IJCAI}, pages 1347--1353.

\bibitem[{Yang et~al.(2016)Yang, Yang, Dyer, He, Smola, and
  Hovy}]{yang2016hierarchical}
Zichao Yang, Diyi Yang, Chris Dyer, Xiaodong He, Alexander~J Smola, and
  Eduard~H Hovy. 2016.
\newblock Hierarchical attention networks for document classification.
\newblock In \emph{HLT-NAACL}, pages 1480--1489.

\bibitem[{Zhang et~al.(2015)Zhang, Zhao, and LeCun}]{zhang2015character}
Xiang Zhang, Junbo Zhao, and Yann LeCun. 2015.
\newblock Character-level convolutional networks for text classification.
\newblock In \emph{Advances in neural information processing systems}, pages
  649--657.

\end{thebibliography}
\bibliographystyle{ijcnlp2017}

\end{document}